\documentclass{article}

\usepackage{microtype}
\usepackage{graphicx}
\usepackage{subcaption}
\usepackage{booktabs} 

\usepackage{hyperref}

\usepackage[preprint]{icml2026}

\usepackage{amsmath}
\usepackage{amssymb}
\usepackage{mathtools}
\usepackage{amsthm}

\usepackage[capitalize,noabbrev]{cleveref}

\theoremstyle{plain}
\newtheorem{theorem}{Theorem}[section]

\theoremstyle{definition}

\newtheorem{assumption}[theorem]{Assumption}
\theoremstyle{remark}
\newtheorem{remark}[theorem]{Remark}

\usepackage[textsize=tiny]{todonotes}
\usepackage{booktabs}
\usepackage{multirow}
\usepackage{placeins}
\usepackage{makecell}

\icmltitlerunning{Personalized Federated Learning for Gradient Alignment}

\begin{document}

\twocolumn[
  \icmltitle{Personalized Federated Learning for Gradient Alignment}

  \icmlsetsymbol{equal}{*}

  \begin{icmlauthorlist}
    \icmlauthor{Dongwon Kim}{comp}
    \icmlauthor{Gyuejeong Lee}{sch}

  \end{icmlauthorlist}

  \icmlaffiliation{comp}{SAKAK Inc., Korea}
  \icmlaffiliation{sch}{Div. of Artificial Intelligence and Data Science, Korea Cyber University, Korea}

  \icmlcorrespondingauthor{Gyuejeong Lee}{lgj413@koreacu.ac.kr}

  \icmlkeywords{Machine Learning, ICML}

  \vskip 0.3in
]

\printAffiliationsAndNotice{}  

\begin{abstract}
Personalized federated learning (pFL) aims to adapt models to client-specific data distributions, yet it often fails to reliably preserve personalized information. Local training is hindered by high-variance gradients induced by limited and heterogeneous client data, while aggregation further distorts client-specific optimization directions. To address these challenges, we propose pFLAlign, a gradient-alignment framework to maintain client-specific information during both local training and aggregation. pFLAlign consists of two complementary mechanisms: one adapts local gradient directions to reduce variance during client-side optimization, and the other mitigates aggregation-induced distortion by realigning the global model with each client’s personalized direction. Theoretically, we derive pFLAlign from a PAC-Bayesian analysis, which reveals how personalized gradient alignment  preserves client-specific information. Our experiments and ablation studies show that pFLAlign consistently improves personalization performance and training stability, achieving state-of-the-art results.

\end{abstract}
\section{Introduction}
As data availability for training large language models (LMs) continues to diminish~\cite{runout}, federated learning (FL)~\cite{fedfed} has emerged as a promising paradigm for distributed model training without data collection, particularly when coupled with small LMs~\cite{openfedllm}. In this setting, personalized federated learning (pFL) has gained attention as it aims to adapt models to client-specific data distributions~\cite{ditto}. Despite its promise, a pivotal challenge in pFL arises from sparse and heterogeneous data distributions across clients, yielding the following optimization issues: (1) Sparse and heterogeneous local data induce noisy gradients that obstruct convergence to personalized optima.~\cite{fedprox,scaffold} (2) Even if clients reach their own optima, aggregation amplifies gradient noise from local updates, pulling client models away from their  optima~\cite{flatre,fedconst}.
\begin{figure}
\vskip 0.2in
\begin{center}
\centerline{\includegraphics[width=\columnwidth]{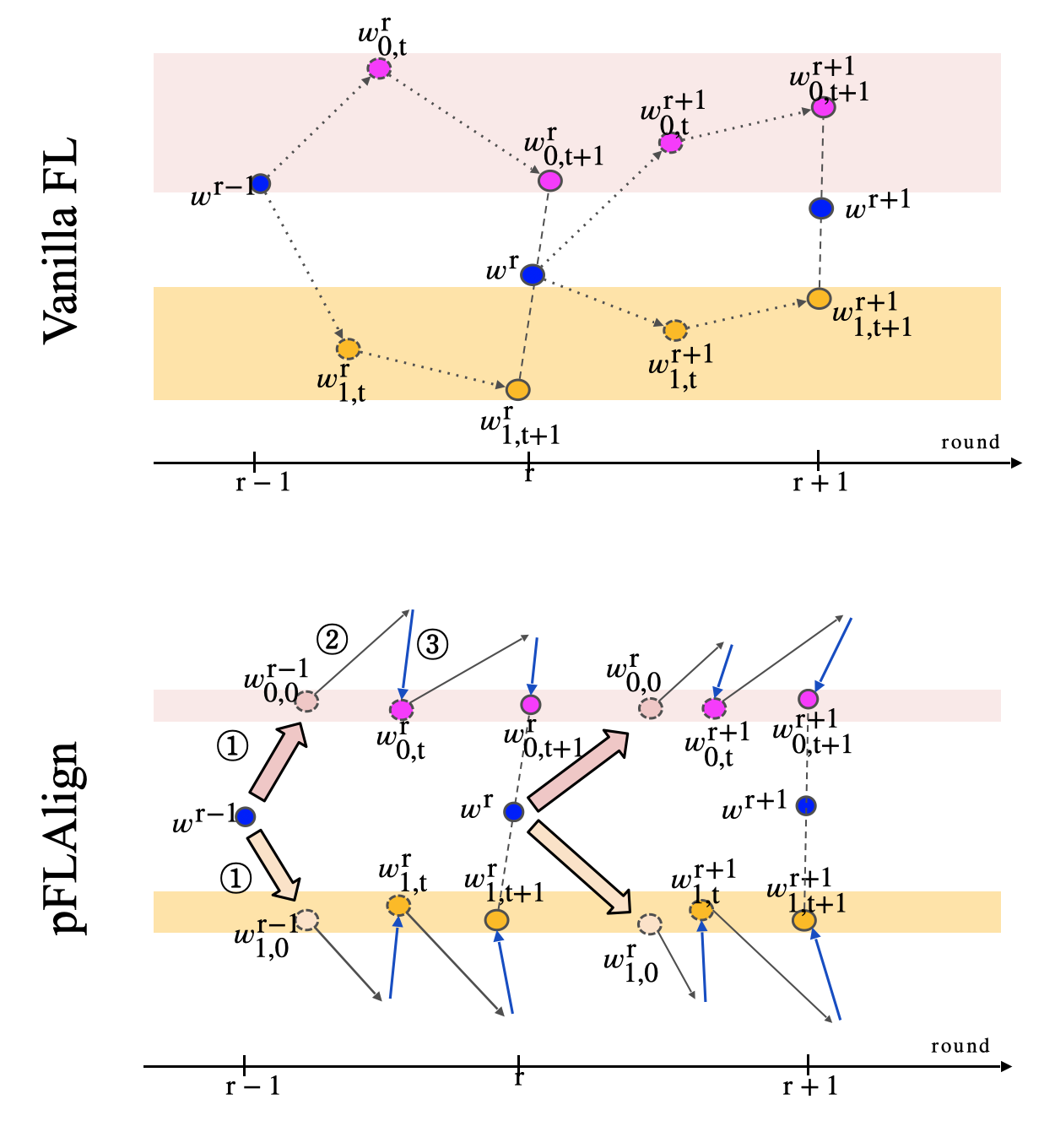}}
\caption{
\textbf{Illustration of the parameter space in FL.}
\emph{Top:} Vanilla FL suffers from high-variance local updates and aggregation-induced distortion,
which drive client models away from personalized optima.
\emph{Bottom:} Our pFLAlign successfully supresses the variance of each client models during local training and aggregation by introducing two complementary mechanisms : Personalized Local Update (2), and Aggregation Robust Personalization (1, 3).
}
\label{fig:concepts}
\end{center}
\vskip -0.2in
\end{figure}

To address noisy gradients of optimization, various FL methods have been proposed to align gradients during optimization. Methods based on correction~\cite{scaffold}, and regularization~\cite{fedprox,fedsam,fedconst} aim to stabilize optimization under heterogeneous client objectives. However, these approaches primarily focus on vision-based settings and global model performance, overlooking the personalization of client models.

Meanwhile, FL studies on LMs have focused on adapter-based personalization and aggregation. FedSA-LoRA~\cite{fedsalora} introduces a dual LoRA architecture that separates local and shared adapters, enabling client-specific parameters to be learned without aggregation. Methods such as FFA-LoRA~\cite{ffa} and FedSVD~\cite{fedsvd} further emphasize aggregation alignment of LoRA adapters by coordinating adapters during aggregation. While controlling gradient noise has been recognized as important in LM training~\cite{adamsecrete, adamheavy, adamtransformer}, the optimization dynamics within FL remain largely unexplored, particularly in pFL.

Here, we raise the central question:

\textit{In pFL, especially for LMs, how can gradient noise be suppressed to preserve personalized information during local updates and the aggregation process?}

To answer this question, we propose \textbf{pFLAlign}, a gradient alignment framework for pFL.
 pFLAlign performs gradient alignment through two complementary mechanisms that target distinct sources of optimization instability. First, pFLAlign employs a \textit{personalized local updates} based on client-specific multiplicative preconditioning, which adaptively scales local gradients to enhance reliable personalized directions and preserve client-specific information across communication rounds. Second, pFLAlign incorporates a \textit{aggregation robust personalization} that combines personalized model initialization with an alignment correction. Personalized initialization biases local optimization toward client-specific optima at the start of each round, while the additive correction term compensates for aggregation-induced distortion by aligning local optimization dynamics with the shared global model. Altogether, these two mechanisms jointly suppress gradient variance arising from both local updates and aggregation, thereby preserving personalized optimization trajectories across communication rounds, as illustrated in Figure~\ref{fig:concepts}.

Conceptually, these two components are inspired by variance-adaptive gradient schemes such as SVAG~\cite{adamdissecting} and variance-reduced correction methods such as SVRG~\cite{svrg}, respectively.

To reinterpret these gradient alignment for pFL, We ground the design of pFLAlign in a PAC-Bayesian analysis that identifies informative parameters under limited and heterogeneous data.  Empirically, our experiments and ablation studies demonstrate that pFLAlign consistently improves personalization performance and training stability, achieving state-of-the-art results across optimization benchmarks and settings.

\begin{itemize}
\item We present a PAC-Bayesian analysis that formally characterizes gradient alignment as a principled mechanism for preserving client-specific information in pFL. To the best of our knowledge, pFLAlign is the first work to provide a PAC-Bayesian treatment of gradient alignment tailored to pFL.

\item Our experiments on heterogeneous datasets and SLM-based FL settings, we demonstrate that pFLAlign achieves stable optimization and state-of-the-art personalization performance.
\end{itemize}

\section{Related Work}

\subsection{Optimization and Federated Learning}
In diverse optimization settings, stochastic gradient methods often suffer from noisy and unstable updates, motivating a rich line of work on variance reductions~\cite{svrg,storm, mars} and adaptive optimizations~\cite{adamw,muon, adamdissecting, choone}. Methods such as SVRG~\cite{svrg} reduce gradient variance by incorporating correction terms based on reference gradients, while adaptive optimizers such as AdamW~\cite{adamw} and related approaches~\cite{muon, adamdissecting} rescale updates using second-order statistics to stabilize optimization under noisy gradients.

These optimizations stabilizing variance have directly influenced the development of FL algorithms, where distributed training and limited local updates further exacerbate instability. To mitigate local drift, FedProx introduces proximal regularization that constrains local updates around the global model. SCAFFOLD~\citep{scaffold} and FedDyn~\citep{feddyn} extend variance-reduction ideas to FL settings by employing correction terms that align local updates with global objectives.

More recent works focus on improving robustness and generalization in federated optimization.
FedSAM~\citep{fedsam} extends sharpness-aware minimization to federated settings, while methods such as FedGLoss~\citep{fedgloss} and FedCONST~\citep{fedconst} regulate update magnitudes or impose convex constraints during aggregation. Although these approaches stabilize global optimization, they do not explicitly address how client-specific optimization directions are preserved under noisy local  and aggregation.

\subsection{Language Model and Federated Learning}

Applying FL to LMs introduces additional challenges due to high-dimensional parameter spaces and noisy gradients~\cite{adamsecrete, adamheavy, adamtransformer}. FedAdamW~\cite{fedadamw} have proposed optimizer-level adaptation to improve convergence and stability for Transformer-based models, providing a PAC-Bayesian analysis.

Adapter-based personalization has also been actively explored in language model–based federated learning. Methods such as FedSA-LoRA~\citep{fedsalora} separate local and shared adapters, while FFA-LoRA~\citep{ffa} and FedSVD~\cite{fedsvd} coordinate LoRA updates to improve aggregation compatibility. Despite strong empirical performance, these approaches primarily focus on architectural design or aggregation alignment, leaving the optimization dynamics of noisy local updates largely unexplored.

\section{Method}
We consider a standard learning setting that trains a deep neural network
$f(x; W)$, where $f : \mathcal{X} \rightarrow \mathcal{Y}$ denotes a neural network composed
of $L$ layers with parameters
\begin{equation}
W = \{ w^{1}, w^{2}, \ldots, w^{L} \}.
\label{eq:network}
\end{equation}

In parameter-efficient training settings such as LoRA~\citep{lora}, the model parameters are decomposed
into a shared backbone and trainable low-rank adapters. Specifically, the parameters are
written as
\begin{equation}
W = W_{0} + \Delta W,
\qquad
\Delta W = \{ \Delta w^{1}, \Delta w^{2}, \ldots, \Delta w^{L} \},
\label{eq:lora_decomposition}
\end{equation}
where $W_{0}$ denotes the shared backbone parameters that remain fixed during training, and
$\Delta W$ represents the trainable low-rank adapter parameters.
Throughout this work, we treat the adapter parameters $\Delta W$ as the effective
learnable model parameters and, for notational simplicity, refer to them as $W$ whenever
the backbone $W_0$ is fixed and implicit. More details are provided in Appendix.

\subsection{Personalized Federated Learning}
We consider a pFL setting, where a shared global model
is trained collaboratively across decentralized client devices while accounting for
client-specific data distributions. Each client $k \in \mathcal{K}$ holds a local dataset
$D_k = \{(x_i, y_i)\}_{i=1}^{N_k}$, where $(x_i, y_i) \sim D_k(x,y)$ are distributed according to a client-specific distribution.

In pFL, the goal is to learn client-specific
adaptations on top of a shared global model. Specifically, we seek a set of personalized
parameters $\Delta W^p =\{\Delta W_k\}_{k \in \mathcal{K}}$  and proper global parameters $W$ as follow:
\begin{equation}
\begin{aligned}
\mathcal{L}(W,\Delta W^p) &= \sum_{k \in \mathcal{K}} \frac{|D_k|}{|D|} \mathcal{L}_k(W,\Delta W_k), \\
\text{where} \quad
\mathcal{L}_k(W,\Delta W_k) &= \mathbb{E}_{(x,y)\sim D_k}\left[\ell(x,y;W+\Delta W_k)\right].
\end{aligned}
\end{equation}

where the global model $W$ captures transferable knowledge across clients, and
$\Delta W_m$ encodes client-specific variations. This formulation decouples global knowledge sharing from personalized adaptation and exposes the central challenge of preserving client-specific information of $\Delta W_m$ during aggregation.

\subsection{pFedAling:Personalized Federated Learning for Gradient Alignment}
Adaptive optimizers such as AdamW~\citep{adamw} and Muon~\citep{muon} achieve strong empirical performance in language model training by reducing gradient noise. A recent study interprets this behavior through the notion of long-term memory, where accumulated second-order statistics stabilize optimization dynamics. Motivated by this observation, we introduce the following conjecture.

\textbf{Conjecture} \textit{Maintaining alignment with client-specific gradient directions during local updates and after aggregation is essential for retaining personalized information.}

Based on this conjecture, we design a pFL framework that enforces gradient alignment throughout FL process to maintain personalization.
More specifically, the design is guided by the following two conditions:

\textbf{Condition 1 (Personalized Local Updates).} \textit{During local optimization, each client’s parameter updates should remain aligned with its personalized optimization direction.} 

\textbf{Condition 2 (Aggregation-Robust Personalization).} \textit{Client models should be aligned with a shared global objective at the client level, especially to stabilize personalized parameters distorted after aggregation.}

To satisfy \textbf{Condition~1},
pFLAlign introduces a \emph{Personalized Local Update} mechanism,
which enforces client-specific gradient alignment during local optimization
through adaptive, multiplicative gradient preconditioning.
 For \textbf{Condition~2},
pFLAlign further incorporates an \emph{Aggregation-Robust Personalization} mechanism.
Specifically, personalized model initialization reorients the global model toward a client-specific direction to mitigate aggregation-induced distortion, while a personalized correction term regularizes local updates exhibiting high variance toward a shared global reference prior to aggregation.

\subsubsection{Gradient Alignment}
Both Condition~1 and Condition~2 aim to enforce gradient alignment under different sources of distortion in FL. To motivate our design, we first revisit how gradient alignment is implicitly handled in existing variance reduction and adaptation methods, and discuss their limitations in preserving personalized optimization directions in pFL.

\textbf{Stochastic Variance Adapted Gradient} Variance adaptation provides an implicit form of gradient alignment by modulating the reliability of each coordinate. In this view, preconditioning determines which gradient directions are trusted during optimization.
To minimize variance-induced error, SVAG optimizer\cite{adamdissecting} introduced the variance adaptation factor on each coordinate $i$ of gradient $g$ introduced by solving:
\begin{equation}
\mathbb{E}\bigl[\|q \odot g - \mathbb{E}[g]\|_2^2 \bigr]
= \sum_{i} q_i^2 \mathbb{E}[g_i^2] - 2q_i\mathbb{E}[g_i] + \mathbb{E}[g_i]^2.
\label{eq:svag}
\end{equation}
Minimizing this gradient deviation yields the optimal preconditioning:
\begin{equation}
q_i = \frac{\mathbb{E}[g_i]^2}{\mathbb{E}[g_i^2]}.
\label{eq:svag2}
\end{equation}
This solution suppresses components dominated by stochastic variance. As a result, SVAG can align gradients toward reliable descent directions. 
In pFL, the expectations $\mathbb{E}[g_i]$ and $\mathbb{E}[g_i^2]$ are no longer defined with respect to a single stationary data distribution, but instead vary across clients and evolve through aggregation. This makes it fail to reflect the reliability of client-specific optimization directions consistently.

\paragraph{Stochastic Variance Reduced Gradient}
Stochastic variance reduced gradient (SVRG)~\cite{svrg} aim to stabilize stochastic optimization by constructing low-variance gradient estimators. Given a reference point $w^{(t)}$ and its full gradient $\nabla f(w^{(t)})$, the SVRG
estimator for a stochastic sample $\xi$ is defined as
\begin{equation}
g_{\text{SVRG}}(w;\xi)
= \nabla f(w;\xi)
- \nabla f(w^{(t)};\xi)
+ \nabla f(w^{(t)}).
\label{eq:svrg}
\end{equation}

In FL, SCAFFOLD adopts this variance-reduction principle by using correction terms that steer client-side gradients toward a shared global optimization direction. However, under heterogeneous data distributions, the reliability of optimization signals varies across clients and evolves over rounds, often in a parameter-specific manner. As a result, indiscriminate global alignment can suppress personalized signals that are critical for certain clients, motivating the need for a more balanced, reliability-aware alignment strategy.

\subsubsection{Personalized Alignment Mechanism}

We introduce two complementary personalized alignment mechanisms:
\emph{Personalized Local Updates} and \emph{Aggregation-Robust Personalization}.
Rather than aligning all client updates to a single global objective, our goal is to ensure that
local training preserves consistent personalized update directions in the presence of noisy gradients ,while client-specific optimization remains stable under aggregation.

\begin{algorithm}[!ht]
\caption{Training procedure of pFLAlign}
\label{alg:pflalign}
\begin{algorithmic}[1]
\REQUIRE Communication rounds $R$, clients $[K]$, local steps $T$, learning rate $\eta$
\ENSURE Global model $w^R$
\STATE Initialize global model $w^0$
\FOR{$r=0,\ldots,R-1$}
    \STATE Sample participating clients $\mathcal{S}^r$
    \FORALL{$k \in \mathcal{S}^r$ \textbf{in parallel}}
        \STATE $w_{k}^{r+1} \leftarrow \text{pFLAlignClient}(k, w^r)$
    \ENDFOR
    \STATE $w^{r+1} \leftarrow \sum_{k\in\mathcal{S}^r} \frac{|D_k|}{\sum_{j\in\mathcal{S}^r}|D_j|} \, w_{k}^{r+1}$
\ENDFOR
\STATE \textbf{Return} $w^R$
\end{algorithmic}

\vspace{0.4em}
\hrule
\vspace{0.4em}

\begin{algorithmic}[1]
\STATE \textbf{\textsc{pFLAlignClient}}$(k, w^r)$:
\STATE \textit{Client state (persistent across rounds):} $\Delta_k^r,\ v,\ P_{k,0}$ \ \ \textit{(init: $\Delta_k^0 \leftarrow 0,\ P_{k,0}\leftarrow 0$)}
\STATE \textit{Client state (reset per round):} $m_0 \leftarrow 0$ 
\STATE \textit{Personalized initialization:}
\STATE $w_{k,0}^r \leftarrow w^r + \Delta_k^r$
\FOR{$t=0,\ldots,T-1$}
    \STATE \textit{Sample minibatch gradient:}
    \STATE $g_{k,t} \leftarrow \nabla f(w_{k,t}^r; \xi_{t}),\ \xi_{t}\sim D_k$
    \STATE \textit{Update personalized preconditioning:}
    \STATE $\begin{aligned}
        &\epsilon \leftarrow 10^{-12} \\
        &m_t \leftarrow \beta m_{t-1} + (1-\beta) g_{k,t},\\
        &v_t \leftarrow \beta v_{t-1} + (1-\beta) g_{k,t}^2,\\
        &\alpha_t \leftarrow 1-(1-\beta)\frac{ g_{k,t}^2}{v_{t}+\epsilon} ,\\
        &\gamma_t \leftarrow 0.5- 0.5\text{erf}(\frac{|m_t|}{\sqrt{2(v_t-m_t^2)+\epsilon}})\text{sign}(-m_t*\Delta_k^r) ,\\
        &P_{k,t} \leftarrow \alpha_t P_{k,t-1}+(1-\beta) \frac{m_{t}^2}{v_t+\epsilon},\\
        &\hat g_{k,t} \leftarrow P_{k,t}\, g_{k,t}
    \end{aligned}$
    \STATE \textit{Alignment correction:}
    \STATE $w_{k,t+1}^r \leftarrow w_{k,t}^r - \eta \hat g_{k,t} - \frac{1}{T}\gamma_t\Delta_k^r$
\ENDFOR
\STATE \textit{Update persistent client state:}
\STATE $\Delta_k^{r+1} \leftarrow (w_{k,T}^r - w^r)$
\STATE \textbf{Return} $w_{k,T}^r$
\end{algorithmic}
\end{algorithm}

\paragraph{Personalized Local Updates } Under noisy and heterogeneous gradients, spersonalization requires local updates to remain consistent with client-specific optimization directions. For this purpose, pFLAlign employs a multiplicative gradient preconditioning scheme that preserves personalized alignment during local training. Motivated by ~\cref{eq:svag2}, we calculate preconditioning term as $\mathbb{E}[g_k]^2/\mathbb{E}[g_k^2]$ , applied independently to each parameter dimension. 
This preconditioner adaptively modulates gradient magnitudes across parameters, suppressing noisy components while preserving personalized directions.
We further introduce a moving factor $\alpha_t$, defined in an element-wise manner, that preserves personalized gradient directions based on parameter sensitivity measured by $g^2_{k,t}$, which reflects minibatch-level loss curvature.
As shown in Section~\ref{sec:analysis}, this preconditioning naturally arises from online PAC-Bayesian framework.
\paragraph{Aggregation-Robust Personalization}
Even when clients converge toward personalized optima, model aggregation can distort client-specific solutions in pFL. We addresses this challenge from a variance-reduction perspective analogous to SVRG and SCAFFOLD, with the key distinction that correction terms are repurposed to stabilize client-specific optimization rather than a shared global objective.
At round $r$, client $k$ starts from a personalized initialization by adding $\Delta_k^{r}$ to the global model. Here, $\Delta_k^{r}$ plays a role analogous to the full gradient $\nabla f(w^{(t)})$ in \cref{eq:svrg}, representing a client-specific direction obtained by approximating the global model gradient on the client’s data.

Although personalized initialization is effective in retaining client-specific information, 
it can also repeatedly amplify noise inherited from previous client updates over global rounds.
To counteract this effect, we apply a correction term $-\gamma_t\frac{\Delta_k^{r}}{T}$ corresponding to the stochastic gradient $-\nabla f(w^{(t)};\xi)$ in \cref{eq:svrg}, which can be interpreted as the global model gradient evaluated on the client’s data. 
Crucially, the strength $-\gamma_t$ of this correction must be carefully controlled to avoid suppressing informative personalized updates.
Recent studies have shown that applying a scaling factor $\gamma_t$ to correction terms is critical for stable optimization,
where the optimal coefficient is often linked to gradient variance or covariance statistics \citep{varco,mars}.

However, reliable estimation of such second-order statistics is challenging due to small and heterogeneous local datasets in FL. Instead, we introduced a probability-based lightweight scaling factor $\gamma_t$ based on the probability that the client's stochastic gradient update aligns with the personalized initialization direction.
Specifically, for each parameter index $i$, we consider probability $\rho_i := \mathbb{P}\big[\text{sign}(-g_{k,i}) = \text{sign}((\Delta_k^r)_i)\big] \text{ on }g_{k,i} \sim \mathcal{N}(m_t, v_t-m_t^2)$ for measuring the probability that the stochastic descent direction agrees with the personalized initialization direction.

To selectively suppress misaligned components,
we scale the correction term by $\gamma_{t,i} = 1 - \rho_i$.
Under this choice, parameters whose gradients are likely to reinforce the personalized direction receive minimal correction,
whereas parameters with inconsistent update directions are strongly attenuated. Assuming the gradient follows a Gaussian distribution
, the alignment probability $\rho_i$ admits a closed-form expression via the error function as in Algorithm~\ref{alg:pflalign}. A detailed discussion of the scaling factor $\gamma_t$ is provided in Section~\ref{sec:analysis}.

\subsection{Theoretic Analysis}
We analyze pFLAlign from a PAC-Bayesian perspective.
First, we study \textit{Personalized Local Updates} through an online PAC-Bayes framework to capture posterior evolution during local training. Similarly, we analyze \textit{Aggregation-Robust Personalization} by viewing the global model as a shared prior. Altogether, we show how our gradient alignment strategy enhance pFL performance via gradient alignment maximizing the utilization of personalized and shared information.
\subsubsection{Online PAC-Bayes Analysis on Personalized Local Updates}
\label{sec:analysis}
In pFL, each client $k$ observes data from its own distribution $\mathcal D_k$, which induces client-specific and sequential optimization dynamics. An online PAC-Bayes framework is particularly suitable, as it naturally accommodates sequential updates and non-IID data streams. Throughout the analysis, we adopt the following standard assumptions.
\begin{assumption}
\label{ass:lsmooth}
(L-smoothness) The empirical risk $\hat R_n(w)$ is $L$-smooth.
\end{assumption}
\begin{assumption}
\label{ass:Bounded Gradient}
(Bounded Gradient) Local optimization follows a preconditioned gradient update $w_{k,t} = w_{k,t-1} - \eta_t \hat{g}_{k,t},$ where $\hat{g}_{k,t}$ is a preconditioned gradient direction satisfying $\|\hat{g}_{k,t}\|\le G$.
\end{assumption}

\begin{assumption}
(Posterior Family) At each local step $t$, the posterior of gradient preconditioner is Gaussian, $Q_{t} = \mathcal N(P_{t}, \Sigma_t),$ centered at $P_{t}$ where covariance is
$\Sigma_t = \mathrm{diag}(s_{t,i}^2)$.
\end{assumption}

Following the online PAC-Bayes framework~\citep{onlinepac} under above assumptions, we associate each local step $t$ of client $k$
with a posterior $Q_{t}$ that recursively depends on previously observed local data.
This view allows us to quantify personalization as controlled deviation of
$Q_{t}$ from the previous posterior $Q_{t-1}$, measured by a KL divergence. 

\begin{theorem}[Online PAC-Bayes Bound for Personalization]
\label{thm:pacbayes-eta}
Fix a client $k$. $R(w)$ is the population risk and $\hat R_{n_k}(w)$ is empirical risk of client data distribution.$n_k$ is the number of client data. Assume that the loss function is bounded in $[0, C]$. For any $\beta>0$ and $\delta\in(0,1)$, with probability at least $1-\delta$
over the local data stream of client $k$, the following holds at each local step $t$:
\begin{align}
\mathbb E_{P_{t}\sim Q_{t}}[R(w_{k,t})]
\;\le\;&
\mathbb E_{P_{t}\sim Q_{t}}[\hat R_{n_k}(w_{k,t})]
\\
&+
\frac{1}{\beta n_k} KL(Q_{t}\|Q_{t-1})
\label{eq:pacbayes}
\nonumber
\\
&+
\frac{\beta C^2}{2}
+
\frac{1}{\beta n_k}\log\frac{1}{\delta}.
\nonumber
\end{align}
where preconditioned gradient update  as $w_{k,t}:=w_{k,t-1} -P_t\odot g_{k,t}$

Minimizing the PAC-Bayes upper bound 
yields the following optimal preconditioning rule on coordinate $i$:
\begin{equation}
p_{t,i}
=
\alpha_{t,i}\,\tilde p_{t,i}
+
(1-\alpha_{t,i})\,p_{t-1,i}
\end{equation}
\[
s_{t,i}^{-2}
=
s_{t-1,i}^{-2}
+
\beta n_k\,L\,G_{t,i}^2,
\]
where
\begin{equation}
\tilde p_{t,i}
=
\frac{\mu_{i}^2}{L\,G_{i}^2},
\qquad
\alpha_{t,i}
=
\frac{L\,G_{i}^2}{
L\,G_{i}^2
+
\frac{1}{\beta n_ks_{t,i}^2}
}
\in(0,1).
\end{equation}

and $\mu = \mathbb E_{(x,y)\sim\mathcal D_k}[\nabla \ell(x,y;w_{k,t})]$.
\end{theorem}
\begin{remark}[Personalization via Controlled Posterior Deviation]
In pFLAlign, personalization arises from allowing each client to maintain a distinct preconditioning posterior $Q$ shaped by its local data distribution $\mathcal{D}_k$. The online PAC-Bayesian bound in Theorem~\ref{thm:pacbayes-eta} characterizes this effect through step-wise KL terms $\mathrm{KL}(Q_t\|Q_{t-1})$, which quantify how local updates adapt the posterior in response to client-specific data. Variance-aware preconditioning naturally emerges as $\tilde p_{t,i}$, corresponding to the ratio $\mathbb{E}[g_i]^2/\mathbb{E}[g_i^2]$ in SVAG.
Moreover, the preconditioning statistics are updated based on curvature-related quantities such as $L G_i^2$, allowing the preconditioner $p_{t,i}$ to retain information from sensitive parameters that exhibit high curvature on a given client minibatch.

\end{remark}

\subsubsection{PAC-Bayes Analysis on Global Objectives}
Personalized initialization introduces client-specific deviations from a shared global model, which may encode useful personalization but also amplify accumulated noise.
To determine how much of this personalized information should be corrected, we require a reference that enables balancing shared and personalized signals in a principled manner.
Accordingly, we define the prior as a distribution of the global model after a single-step update based on client gradient statistics, and the posterior as the client model after personalized initialization.
\begin{theorem}[PAC-Bayes Bound for Aggregation-Robust Personalization]
\label{thm:pacbayes-global}
Fix a client $k$. Let $R(w)$ denote the population risk and $\hat R_{n_k}(w)$ the empirical risk
on client $k$'s local data. $n_k$ is the number of client data.
Assume the loss is bounded in $[0,C]$.
For any prior distribution $\pi$,
any posterior distribution $Q$, any $\lambda>0$, and any $\delta\in(0,1)$,
with probability at least $1-\delta$ over the local dataset of client $k$,
the following holds:
\[
\mathbb{E}_{w\sim Q}[R(w)]
\le
\mathbb{E}_{w\sim Q}[\hat R_{n_k}(w)]
+
\frac{\mathrm{KL}(Q\|\pi)+\log\frac{1}{\delta}}{\lambda}
+
\frac{\lambda C^2}{8n_k}.
\]
Consider the prior $\pi$ as the one-step stochastic-gradient update distribution
from the global reference $w^r$ on client $k$,$\text{ resulting prior }\pi = \mathcal N\!\big(w^r-\eta m_k,\; \eta^2\Sigma_k\big).$  where the gradient variance is \( \Sigma_k = \mathrm{diag}(\rho_i^2) \). Let the posterior be point of mass at the personalized initialization, following the Dirac distribution $Q = \delta_{w^r+\Delta_k^r}$. Then, the KL-gradient with respect to the personalized offset is
\[
\big[\nabla_{\Delta_k^r}\mathrm{KL}(Q\|\pi)\big]_i
=
\frac{(\Delta_k^r)_i + \eta\, m_{k,i}}{\eta^2\,\rho_i^2}.
\]
In particular, when $\eta |m_{k,i}|\ll |\Delta_{k,i}^r|$,
the KL gradient is dominated by a variance-weighted shrinkage term proportional to
$(\Delta_k^r)_i$. Moreover, if the descent direction $-\eta m_{k,i}$ is misaligned with
the personalized offset $(\Delta_k^r)_i$, this shrinkage acts to strongly penalize
the deviation from the global reference, whereas aligned directions incur a weaker penalty.

\end{theorem}

\begin{remark}[KL Control via Probabilistic Alignment]
\label{prop:kl-control}
The shrinkage term in Theorem~\ref{thm:pacbayes-global} shows how penalizing deviations
of the personalized offset $\Delta_k^r$ controls the PAC-Bayesian complexity. However, directly  applying this shrinkage may destabilize the optimization process. To modify KL-induced shrinkage stable, we modulate it in a probabilistic manner based on alignment with the current descent direction. Modeling the stochastic gradient as a Gaussian distribution with mean $m_k$,
the probability that the descent direction $-\eta m_{k,i}$ at index $i$ disagrees with
the personalized offset $(\Delta_k^r)_i$ admits a closed-form expression via the Gaussian error function. This yields the scaling factor
$\gamma_{t,i} \propto \big[\nabla_{\Delta_k^r}\mathrm{KL}(Q\|\pi)\big]_i$,
which acts as a probabilistic gate on the KL-induced shrinkage in the regime
$|\eta m_{k,i}|\ll |\Delta_{k,i}^r|$. Such a regime naturally arises, as $\Delta_k^r$ aggregates client-specific update directions collected over multiple optimization steps with reduced variance, whereas $\eta m_{k,i}$ reflects a single-step update estimated with stochastic noise.

\end{remark}

\begin{table*}[!ht]
\centering
\small
\setlength{\tabcolsep}{4.0pt}
\renewcommand{\arraystretch}{1.15}
\caption{\textbf{Personalized FL performance under task-level heterogeneity.}
ROUGE-L (\%) over three seeds with 95\% CIs (Student’s $t$).
Baselines include LoRA-modulation (FFA-LoRA\citep{ffa}, FedSA-LoRA\citep{fedsalora}, Fed-SVD\citep{fedsvd})
and optimizer-based methods (FedAvg, FedProx\citep{fedprox}, FedSAM\citep{fedsam}, SCAFFOLD\citep{scaffold}, FedDyn\citep{feddyn}, FedYogi).
Client task assignment ($c_1$--$c_4$):
FLAN—coreference, entailment, paraphrase, structure-to-text;
Dolly—closed QA, information extraction, classification, summarization.
}

\label{tab:clientwise_grouped_final}

\begin{tabular}{ll|cccc||c|cccc||c}
\toprule
\multirow{2}{*}{\textbf{Model}} &
\multirow{2}{*}{\textbf{Algorithm}} &
\multicolumn{5}{c|}{\textbf{ Databricks-dolly-15k}} &
\multicolumn{5}{c}{\textbf{FLAN}} \\
\cmidrule(lr){3-7}\cmidrule(lr){8-12}
& &
\textbf{C1} & \textbf{C2} & \textbf{C3} & \textbf{C4} & \textbf{Avg} &
\textbf{C1} & \textbf{C2} & \textbf{C3} & \textbf{C4} & \textbf{Avg} \\
\midrule

\multirow{9}{*}{\textbf{\makecell[l]{\textbf{LLaMA-3.2}\\\textbf{-Instruct-1B}}}} &
\multicolumn{11}{l}{\textit{LoRA-modulation methods}} \\
\cmidrule(lr){2-12}
& FFA-LoRA     
& 40.32 & 34.33 & 39.73 & 36.31 & 37.68$\pm$0.92 
& 63.04 & 31.71 & 21.88 & 44.61 & \underline{40.31}$\pm$3.81 \\

& FedSA-LoRA   
& \textbf{47.61} & 33.53 & \textbf{42.71} & 34.73 & \textbf{39.65}$\pm$0.89 
& 41.51 & 6.30 & 3.52 & 42.72 & 23.51$\pm$8.69 \\

& Fed-SVD
& 39.93 & 32.11 & 39.96 & 35.16 & 36.79$\pm$2.13 
& 60.45 & 26.49 & 14.83 & \textbf{45.04} & 34.79$\pm$5.11 \\

\addlinespace[2pt]
& \multicolumn{11}{l}{\textit{Optimizer-based methods}} \\
\cmidrule(lr){2-12}
& FedAvg       
& 38.93 & 33.18 & 41.00 & 34.81 & 36.98$\pm$2.37
& 58.32 & 28.40 & 17.61 & 43.49 & 36.96$\pm$3.14 \\

& FedProx      
& 40.29 & 32.16 & 41.42 & 36.23 & 37.52$\pm$4.13
& 59.93 & 28.72 & 13.61 & 42.68 & 36.15$\pm$3.45 \\

& FedSAM       
& 37.76 & 29.82 & 37.10 & 32.61 & 34.32$\pm$0.84
& 49.61 & 12.12 & 2.45 & 35.38 & 24.89$\pm$5.11 \\

& SCAFFOLD     
& 38.72 & 30.42 & 41.90 & 35.25 & 36.57$\pm$3.35
& 57.40 & 32.98 & 17.64 & 44.05 & 38.02$\pm$3.38 \\

& FedDyn       
& 39.42 & \textbf{34.60} & 39.32 & 35.64 & 37.24$\pm$1.79
& 60.20 & 28.77 & 13.18 & 41.86 & 36.00$\pm$3.10 \\

& FedYoGI      
& 39.20 & 34.26 & 37.68 & 34.12 & 36.31$\pm$1.86
& 58.28 & 13.81 & 5.06 & 38.51 & 28.91$\pm$4.69 \\

& \textbf{(Ours)} 
&40.19 & 33.46 & 41.58 & \textbf{35.03} & \underline{37.57}$\pm$2.66
& \textbf{63.73} & \textbf{34.90} & \textbf{40.64} & 37.97 & \textbf{44.31}$\pm$5.19 \\
\midrule

\multirow{9}{*}{\textbf{OPT-1.3B}} &
\multicolumn{11}{l}{\textit{LoRA-modulation methods}} \\
\cmidrule(lr){2-12}
& FFA-LoRA     
& 49.68 & 34.16 & 31.14 & 36.67 & 37.91$\pm$0.60
& 61.09 & 30.00 & 47.33 & 42.86 & 45.32$\pm$5.40 \\

& FedSA-LoRA   
& 49.36 & 35.46 & 32.00 & 37.38 & \underline{38.55}$\pm$1.15
& 62.98 & 31.33 & 49.00 & 47.19 & 47.62$\pm$3.60 \\

& Fed-SVD      
& 48.41 & 34.89 & 31.95 & 36.89 & 38.04$\pm$0.85
& 63.05 & 32.33 & 45.83 & 43.43 & 46.16$\pm$5.25 \\

\addlinespace[2pt]
& \multicolumn{11}{l}{\textit{Optimizer-based methods}} \\
\cmidrule(lr){2-12}
& FedAvg       
& 47.88 & 33.34 & 30.97 & 35.94 & 37.04$\pm$1.45
& \textbf{63.79} & 31.50 & 50.67 & 42.15 & 47.03$\pm$7.55 \\

& FedSAM       
& 50.26 & 34.17 & 31.42 & 36.29 & 38.01$\pm$0.12
& 61.57 & 35.33 & 35.33 & 41.50 & 43.44$\pm$0.95 \\

& FedGloss     
& 23.19 & 17.19 & 18.01 & 27.26 & 21.41$\pm$8.70
& 9.05  & 0.97  & 1.39  & 10.96 & 5.59$\pm$10.12 \\

& FedYoGI      
& 48.34 & 33.18 & 29.96 & 35.60 & 36.77$\pm$0.15
& 63.57 & 33.00 & \textbf{61.00} & 41.20 & \underline{49.69}$\pm$2.29 \\

& SCAFFOLD     
& 47.93 & 35.10 & 32.12 & 35.94 & 37.77$\pm$0.50
& 63.78 & 31.33 & 50.00 & 43.88 & 47.25$\pm$2.78 \\

& FedDyn       
& 48.51 & 33.58 & 31.00 & 35.23 & 37.08$\pm$0.30
& 62.36 & 31.33 & 57.50 & 42.63 & 48.46$\pm$3.06 \\

& \textbf{(Ours)}
& \textbf{51.51} & \textbf{38.15} & \textbf{36.36} & \textbf{37.20} & \textbf{40.81}$\pm$3.01
& 62.81 & 32.17 & 58.00 & \textbf{47.67} & \textbf{50.17}$\pm$7.6 \\

\bottomrule
\end{tabular}
\end{table*}
\section{Experiments}

\subsection{Experimental Settings}
\textbf{Models and Datasets.}
We use LLaMA-3.2~\citep{llama3} and OPT-1.3B~\citep{opt} to conduct federated instruction tuning on Databricks-Dolly-15K~\citep{dolly} (with contextual information) and FLAN~\citep{flan} (without contextual information).
For each data setting, a disjoint set of four tasks is randomly assigned to each client to induce task-level heterogeneity.

\textbf{Baselines.}
We categorize baselines into \emph{LoRA-modulation methods} and \emph{optimizer-based methods}.
The former focuses on architectural or parameterization-level adaptations of LoRA modules,
while the latter modifies optimization dynamics to improve training stability and personalization under federated settings.

\textbf{FL Setting.}
We follow a synchronous FL protocol with 4 clients over 50 communication rounds, where all clients participate in every round.
At each round, clients perform 5 local optimization steps with a batch size of 4 and a sequence length of 1024.
We employ LoRA for parameter-efficient tuning with rank 16 and scaling factor 32, and use a fixed learning rate of $4\times10^{-2}$ for all experiments.
Model performance is evaluated using ROUGE-L.
Additional details are provided in the Appendix.

\subsection{Performance Comparison}
Dolly provides dense contextual information but remains challenging due to its complexity, leading to lower performance across all methods, particularly for the smaller LLaMA model. In contrast, FLAN contains sparse task information without context, which implicitly induces severe task-level heterogeneity across clients. Under this regime, several baselines suffer from significant client-wise performance imbalance. Notably, FedSA-LoRA exhibits substantial performance degradation, highlighting its reliance on dense task information for stable adaptation.

Our method achieves the best average performance across all settings, except for Dolly with LLaMA, where it ranks second.
We attribute this gap to the limited adaptation capacity of low-rank LoRA modules in small models rather than a limitation of our optimization framework. This interpretation is supported by the strong performance of our method on OPT under the same dataset setting, where increased model capacity enables more effective utilization of personalized updates.

\subsection{Alignment on Personalized Objectives}
\begin{figure}[h]
\vskip -0.0in
\begin{center}
\centerline{\includegraphics[width=1.0\columnwidth]{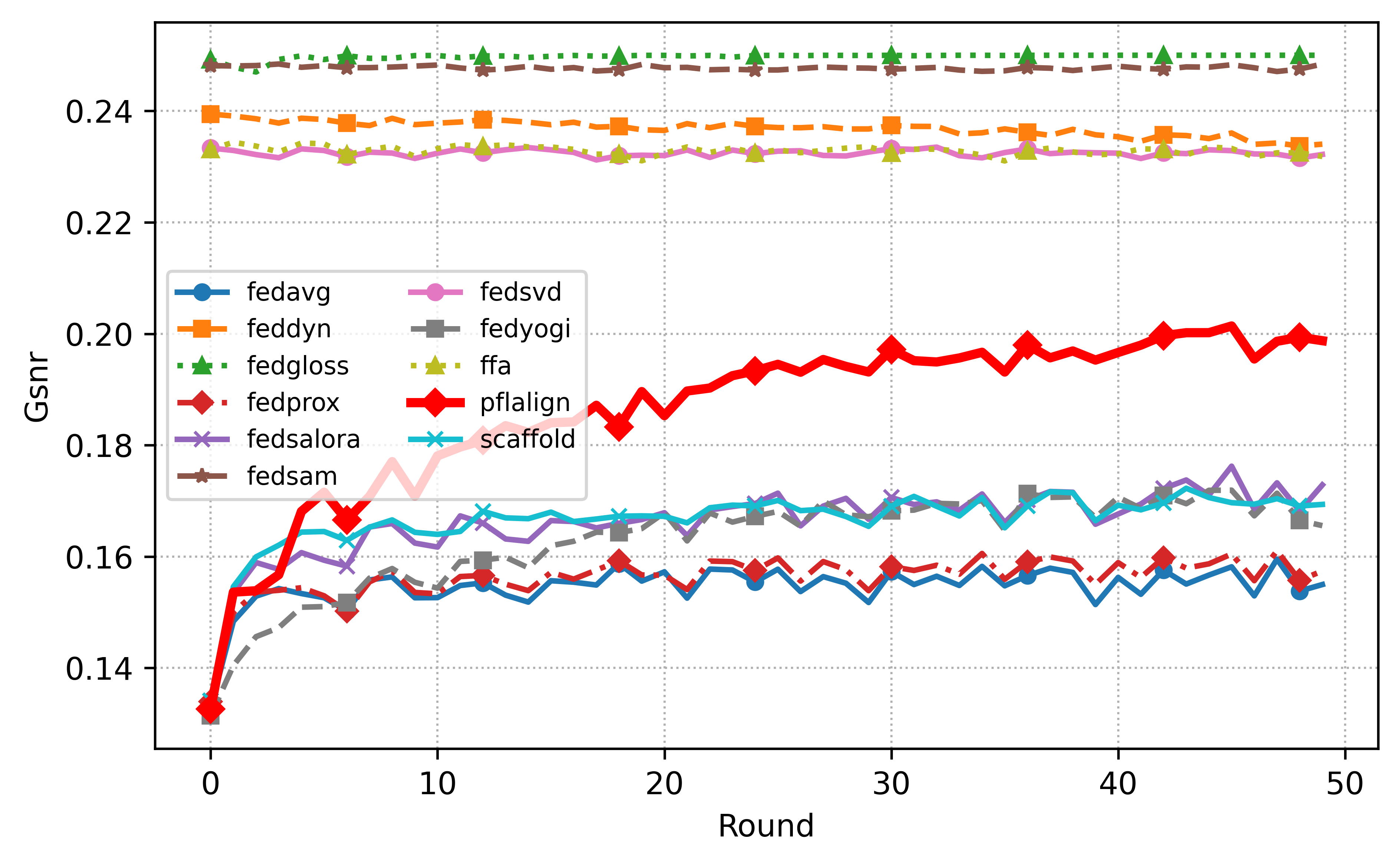}}
\caption{\textbf{GSNR value on a Client}Gradient signal-to-noise ratio (GSNR) measured on a single client for the structure-to-text task from the FLAN dataset. } 
\label{fig:abgsnr}
\end{center}
\vskip -0.2in
\end{figure}
\paragraph{Personalized Direction Alignment}To analyze how different methods align client updates with personalized objectives, we examine the \emph{gradient signal-to-noise ratio (GSNR)}. GSNR measures the ratio between the magnitude of the mean gradient signal and the variance of stochastic gradient updates, thereby serving as an indicator of optimization stability and directional consistency. Figure~\ref{fig:abgsnr} illustrates the GSNR trajectory measured on Client~3. Compared to existing baselines, \textbf{pFLAlign}  consistently exhibits the steepest increase in GSNR throughout training. This behavior indicates that, as optimization progresses, client gradients under pFLAlign become increasingly aligned with the personalized objective direction while suppressing stochastic noise. In contrast, other methods either exhibit slower growth on GSNR trends, suggesting weaker alignment and higher variance in local update directions.

\begin{figure}[h]
\begin{center}
\centerline{\includegraphics[width=1.0\columnwidth]{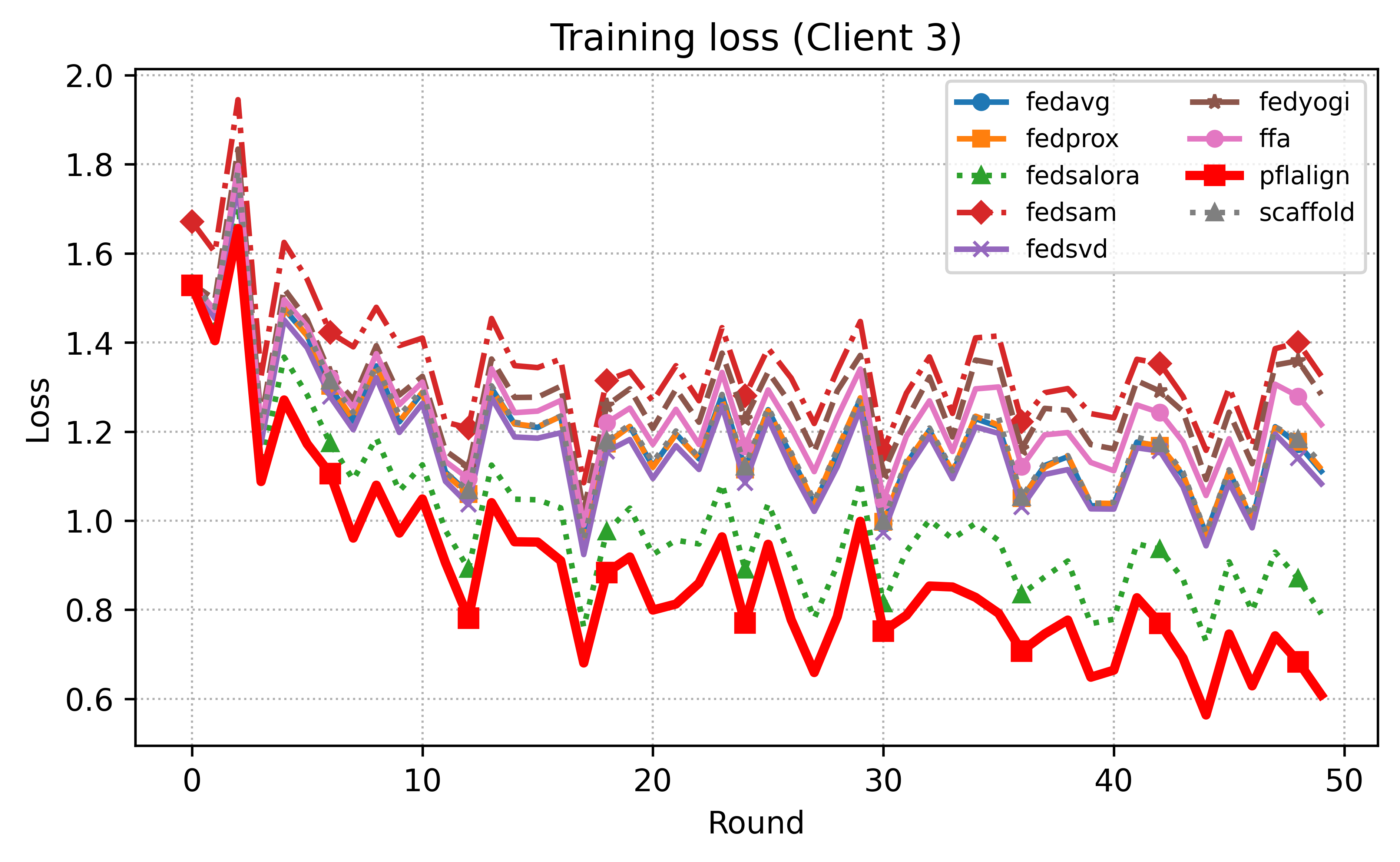}}
\caption{\textbf{Loss value on a Client} Training loss curves on a single client trained on the structure-to-text task from the FLAN dataset.} 
\label{fig:loss}
\end{center}
\vskip -0.2in
\end{figure}
\paragraph{Personalized Objective Alignments}We further examine the training loss to assess the practical impact of personalized alignment.
Figure~\ref{fig:loss} reports the training loss on Client~3 under the same random seed.
Across the entire training trajectory, \textbf{pFLAlign} consistently achieves lower loss values compared to competing methods. This sustained loss reduction suggests that the improved GSNR directly translates into more effective optimization toward the personalized objective.

\begin{figure}[h]
\vskip 0.2in
\begin{center}
\centerline{\includegraphics[width=1.0\columnwidth]{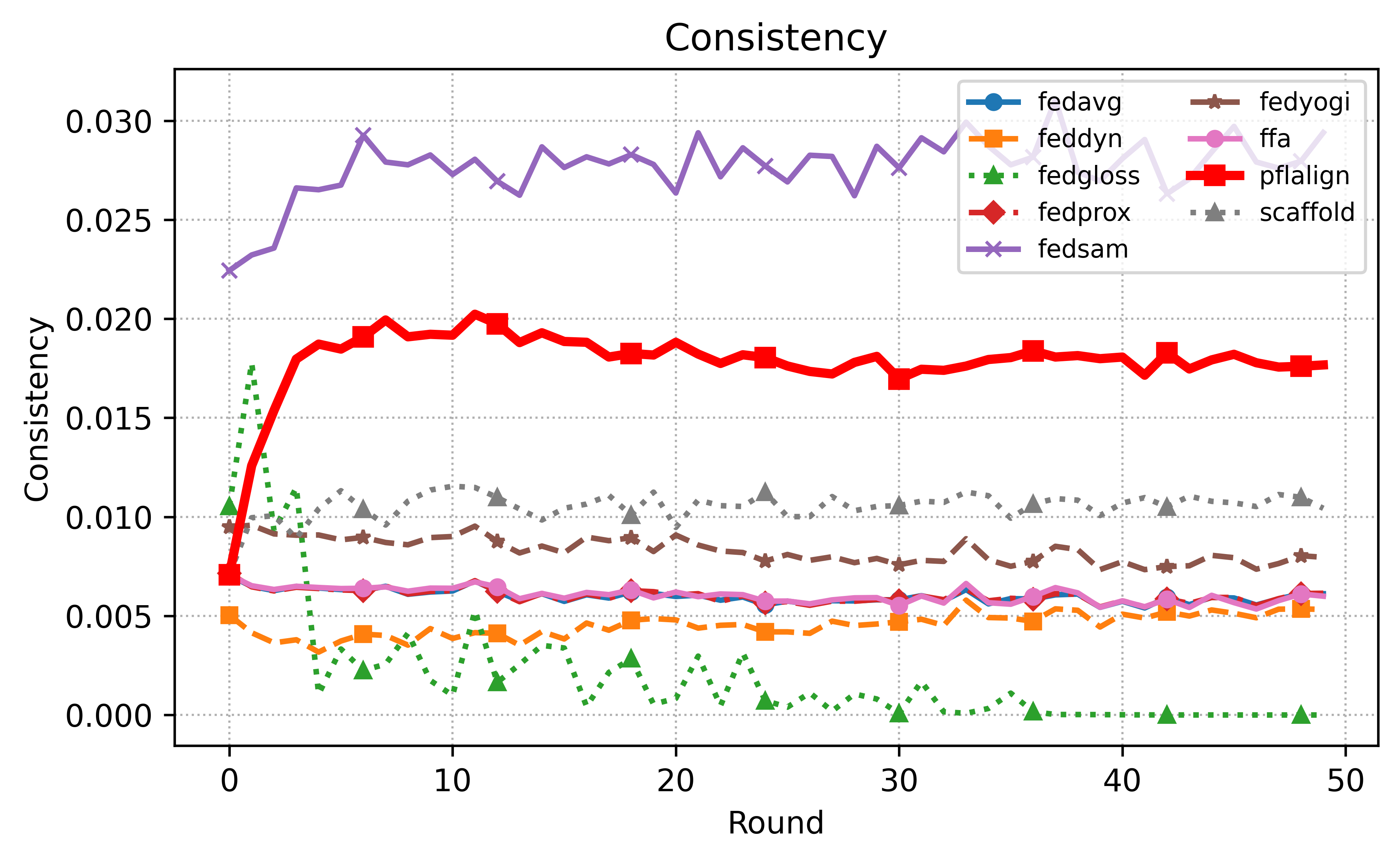}}
\caption{\textbf{Aggregation Consistency.}
Consistency of aggregated updates across rounds, measured by the data-weighted norm of client updates.
}

\label{fig:abg}
\end{center}
\vskip -0.2in
\end{figure}

\paragraph{Aggregation Consistency.}
Aggregation consistency is defined as the data-weighted norm of client updates,
\(
\sum_{k\in\mathcal{S}^r} \frac{|D_k|}{\sum_{j\in\mathcal{S}^r}|D_j|} \|w_k^{r+1}\|,
\)
which measures how much client-specific information is reflected in the aggregated model.
Figure~\ref{fig:abg} reports the aggregation consistency across different methods.
We observe that personalized initialization substantially increases consistency, indicating that a larger portion of client-specific update information is effectively incorporated into the aggregated model.
Notably, while SAM also exhibits high consistency, GSNR analysis reveals that such alignment is largely driven by noise-induced effects.
In contrast, pFLAlign maintains high consistency together with higher GSNR, indicating that client information is preserved through reliable and well-aligned gradient signals.

\section{Conclusion}
In this work, we investigated pFL through the lens of gradient alignment under noisy and heterogeneous optimization, with a focus on LM fine-tuning. Personalization is fundamentally hindered by both stochastic local gradients and aggregation-induced distortion of client-specific optimization directions. To address these, we proposed pFLAlign, a gradient alignment framework that preserves personalized information during both local training and aggregation via personalized preconditioning and aggregation-robust alignment correction.
We ground the design of pFLAlign in a PAC-Bayesian analysis, which provides principled guarantees on preserving informative personalized directions .
Experiments on federated instruction-tuning benchmarks demonstrate that pFLAlign consistently improves personalization performance and training stability of optimization. Our results highlight gradient alignment as a key mechanism for effective pFL, particularly on LM settings.

\bibliography{main}
\bibliographystyle{icml2026}

\newpage
\appendix
\onecolumn

\section{Client-wise Task Assignment and Data Statistics}
\label{sec:appendix-data}

We provide detailed statistics of the client-wise task assignment and data distribution used in our federated learning experiments.
To induce task-level heterogeneity, each client is assigned a single task, and the client--task mapping is fixed throughout all communication rounds and random seeds.
Training and test sets are constructed separately for each client according to its assigned task.

For the FLAN\citep{flan} dataset, all tasks are assigned the same number of training and test samples.
For Databricks-Dolly-15K\citep{dolly}, the number of samples varies across tasks following the original dataset composition.
Table~\ref{tab:data_stats} summarizes the number of training and test samples assigned to each client.
\begin{table}[h]
\centering
\caption{\textbf{Client-wise task assignment and data statistics.}
Each client is assigned a single task.
We report the number of training and test samples for each client and dataset.}
\label{tab:data_stats}
\begin{small}
\begin{tabular}{c|c|c|c|c}
\hline
\textbf{Dataset} & \textbf{Client} & \textbf{Task} & \textbf{\# Train} & \textbf{\# Test} \\
\hline
\multirow{4}{*}{FLAN}
& $c_1$ & Coreference        & 300 & 200 \\
& $c_2$ & Entailment         & 300 & 200 \\
& $c_3$ & Paraphrase         & 300 & 200 \\
& $c_4$ & Structure-to-Text  & 300 & 200 \\
\hline
\multirow{4}{*}{Databricks-Dolly-15K}
& $c_1$ & Classification            & 1919 & 214 \\
& $c_2$ & Closed QA                 & 1608 & 178 \\
& $c_3$ & Information Extraction    & 1919 & 151 \\
& $c_4$ & Summarization             & 1052 & 119 \\
\hline
\end{tabular}
\end{small}
\end{table}

\section{Online PAC-Bayes Update for Parameter-wise Preconditioner \textbf{$p_{t,i}$}}

For a sequence of posteriors $\{Q_t\}$ of preconditioners, the online PAC-Bayes bound yields
\[
\mathbb{E}_{W \sim Q_t}[R(W)]
\le
\mathbb{E}_{W \sim Q_t}[\hat{R}_n(W)]
+
\frac{1}{\beta n}
\Bigl(
KL(Q_t \,\|\, Q_{t-1}) + \log \tfrac{1}{\delta}
\Bigr)
+
\frac{\beta K^2}{2}.
\]

From Assumption~\ref{ass:lsmooth} with the preconditioned update
$W' = W - Pg$, we have
\[
\hat{R}_n(W - Pg)
\le
\hat{R}_n(W)
- \langle \mu, Pg\rangle
+ \frac{L}{2}\|Pg\|^2 .
\]

Expanding the inner products coordinate-wise gives
\[
\langle \mu, Pg\rangle = \sum_i \mu_i (p_i g_i),
\qquad
\|Pg\|^2 = \sum_i p_i^2 g_i^2 .
\]

We model each coordinate-wise preconditioner as a random variable and take expectation with respect to both the gradient randomness and the posterior distribution over $p_i$.
This yields
\[
\mathbb{E}[\hat{R}_n(W - P\odot g)]
\le
\mathbb{E}[\hat{R}_n(W)]
-
\sum_i \mathbb{E}_{Q_i}[p_i]\,\mu_i^2
+
\frac{L}{2}
\sum_i \mathbb{E}_{Q_i}[p_i^2]\,G_i^2
+
\frac{1}{\beta n} KL(Q\|P)
+
\frac{\beta K^2}{2}.
\]

We assume a Gaussian posterior for each coordinate-wise preconditioner,
\[
Q_i = \mathcal{N}(m_{t,i}, s_{t,i}^2),
\qquad
P_i = \mathcal{N}(m_{t-1,i}, \tau_i^2),
\]
where $Q_{t-1}$ serves as the prior at round $t$.
Under this assumption,
\[
\mathbb{E}_{Q_i}[p_i] = m_{t,i},
\qquad
\mathbb{E}_{Q_i}[p_i^2] = m_{t,i}^2 + s_{t,i}^2.
\]
Moreover, recall that the posterior and prior distributions for coordinate $i$ are defined as
\[
Q_i = \mathcal N(p_{t,i}, s_{t,i}^2),
\qquad
P_i = \mathcal N(p_{t-1,i}, s_{t-1,i}^2),
\]
respectively. For univariate Gaussians, the KL divergence admits the closed form
\[
\mathrm{KL}(Q_i \| P_i)
=
\frac{1}{2}
\left(
\frac{(p_{t,i}-p_{t-1,i})^2 + s_{t,i}^2}{s_{t-1,i}^2}
+
\log\frac{s_{t-1,i}^2}{s_{t,i}^2}
-1
\right).
\]

\paragraph{Coordinate-wise objective.}
Substituting the above expression into the PAC-Bayesian bound and collecting all coordinate-dependent terms yields the following objective (up to constants):
\[
J_i(p_{t,i}, s_{t,i}^2)
=
- p_{t,i} \mu_i^2
+ \frac{L}{2} G_i^2 \bigl(p_{t,i}^2 + s_{t,i}^2\bigr)
+ \frac{1}{\beta n} \mathrm{KL}(Q_i \| P_i).
\]
Here, the first two terms arise from a second-order upper bound on the empirical risk under the $L$-smoothness assumption, where $\mu_i^2$ and $G_i^2$ denote the mean and second moment of the stochastic gradient along coordinate $i$, respectively.

\paragraph{Posterior mean update.}
Taking the derivative of $J_i$ with respect to $p_{t,i}$ and setting it to zero yields
\begin{equation}
\label{eq:mean_update}
p_{t,i}
=
\frac{
\mu_i^2
+
\frac{1}{\beta n}\frac{p_{t-1,i}}{s_{t-1,i}^2}
}{
L G_i^2
+
\frac{1}{\beta n}\frac{1}{s_{t-1,i}^2}
}.
\end{equation}
This update can be interpreted as a precision-weighted average between the current gradient signal and the prior mean.

\paragraph{Posterior variance update.}
Minimizing $J_i$ with respect to $s_{t,i}^2$ yields
\begin{equation}
\label{eq:var_update}
s_{t,i}^2
=
\frac{1}{
\frac{1}{\tau_i^2}
+
\beta n\,L\,G_i^2
}.
\end{equation}
Identifying the prior variance as $\tau_i^2 = s_{t-1,i}^2$, the update can equivalently be expressed in terms of the posterior precision as
\[
s_{t,i}^{-2}
=
s_{t-1,i}^{-2}
+
\beta n\,L\,G_{t,i}^2.
\]
This form highlights that curvature information accumulates additively over rounds, indicating that uncertainty in the preconditioner is monotonically reduced by accumulated curvature information.
\paragraph{Weighted-average interpretation.}
The posterior mean update in~\eqref{eq:mean_update} can be equivalently rewritten as a convex combination
\[
p_{t,i}
=
\alpha_{t,i}\,\tilde p_{t,i}
+
(1-\alpha_{t,i})\,p_{t-1,i},
\]
where
\[
\tilde p_{t,i}
=
\frac{\mu_{i}^2}{L\,G_{i}^2},
\qquad
\alpha_{t,i}
=
\frac{L\,G_{i}^2}{
L\,G_{i}^2
+
\frac{1}{\beta n}\,s_{t-1,i}^{-2}
}
\in (0,1).
\]

Here, $s_{t-1,i}^2$ denotes the posterior variance of the preconditioner from the previous round.
Recall that the posterior precision follows a cumulative update
\[
s_{t-1,i}^{-2}
=
\tau_i^{-2}
+
\sum_{k < t} \beta n\,L\,G_{i}^2,
\]
indicating that the denominator of $\alpha_{t,i}$ aggregates curvature information from all past rounds.

Substituting the cumulative precision explicitly, the mixing coefficient can be written as
\[
\alpha_{t,i}
=
\frac{L\,G_{t,i}^2}{
L\,G_{t,i}^2
+
\sum_{k < t} \,L\,G_{k,i}^2
}
\]

This expression makes explicit that $\alpha_{t,i}$ compares the curvature information contributed by the current round to the total curvature precision accumulated from all previous rounds.
As the sum $\sum_{k < t} G_{k,i}^2$ grows, the weight $\alpha_{t,i}$ naturally decreases, reflecting increasing confidence in the preconditioner, whereas in early rounds or under distributional shifts, newly observed gradient
energy dominates and $\alpha_{t,i}$ becomes large.

In practice, the curvature proxy $G_{m,i}^2$ is approximated by the squared stochastic gradient, i.e.,
\[
LG_{i}^2 \;\approx\; g_{i}^2,
\]
so that $\alpha_{t,i}$ directly reflects the relative contribution of the current gradient energy to the accumulated curvature precision.
Consequently, the update weight $\alpha_{t,i}$ becomes large when new gradient information dominates prior uncertainty, and small when the preconditioner has already converged with high confidence.

\paragraph{Final update.}
The resulting PAC-Bayesian parameter-wise preconditioning rule is
\[
W_{t+1} = W_t - P_t g_t,
\qquad
P_t = \mathrm{diag}(p_{t,1},\dots,p_{t,d}),
\]
where the variance term $s_{t,i}^2$ directly regularizes the curvature penalty, ensuring that uncertain coordinates are updated conservatively.

\section{Proof of Theorem~\ref{thm:pacbayes-global}}
\label{app:pacbayes-global}

We provide a derivation of Theorem~\ref{thm:pacbayes-global} by instantiating a standard PAC-Bayes bound with a data-dependent prior induced by a one-step stochastic gradient update.

\paragraph{PAC-Bayes bound.}
For a fixed client $k$, consider a loss function bounded in $[0,C]$ and a local dataset of size $n_k$.
For any prior distribution $\pi$, posterior distribution $Q$, and $\lambda>0$, with probability at least $1-\delta$ over the draw of the dataset,
\begin{equation}
\label{eq:pacbayes}
\mathbb{E}_{w\sim Q}[R(w)]
\le
\mathbb{E}_{w\sim Q}[\hat R_{n_k}(w)]
+
\frac{\mathrm{KL}(Q\|\pi)+\log\frac{1}{\delta}}{\lambda}
+
\frac{\lambda C^2}{8n_k}.
\end{equation}

Let $w^r$ denote the global reference model at communication round $r$.
We define the prior $\pi$ as the distribution induced by a single stochastic gradient step on client $k$,
\begin{equation}
\pi = \mathcal{N}\!\big(w^r - \eta m_k,\; \eta^2 \Sigma_k\big),
\end{equation}
where $m_k = \mathbb{E}[g_k]$ denotes the mean local gradient and $\Sigma_k = \mathrm{diag}(\rho_i^2)$
represents a diagonal approximation of gradient variance.

The posterior $Q$ is chosen as a Dirac distribution centered at the personalized initialization,
\begin{equation}
Q = \delta_{w^r + \Delta_k^r}.
\end{equation}

Since $Q$ is a Dirac distribution, the KL divergence reduces to the negative log-density of $\pi$ evaluated at $w^r+\Delta_k^r$ up to an additive constant:
\begin{equation}
\mathrm{KL}(Q\|\pi)
=
\frac{1}{2}
\sum_i
\frac{\big((\Delta_k^r)_i + \eta m_{k,i}\big)^2}{\eta^2 \rho_i^2}
+ \text{const}.
\end{equation}

Taking the gradient with respect to $\Delta_k^r$ yields
\begin{equation}
\label{eq:kl-grad}
\big[\nabla_{\Delta_k^r}\mathrm{KL}(Q\|\pi)\big]_i
=
\frac{(\Delta_k^r)_i + \eta m_{k,i}}{\eta^2 \rho_i^2}.
\end{equation}

When $|\eta m_{k,i}|\ll |(\Delta_k^r)_i|$, the KL gradient is dominated by a variance-weighted shrinkage term proportional to $(\Delta_k^r)_i$, which penalizes large deviations from the global reference.
Moreover, if the descent direction $-\eta m_{k,i}$ is misaligned with $(\Delta_k^r)_i$, the effective shrinkage is amplified, whereas aligned directions incur a weaker penalty.
This behavior motivates the probabilistic modulation of KL-induced shrinkage introduced in the main text.

\section{Derivation of Alignment Probability}
\label{app:alignment}

We derive the closed-form expression for the alignment probability $\rho_i$ used to construct the scaling factor $\gamma_{t,i}$.

For each parameter index $i$, let the stochastic gradient be modeled as
\begin{equation}
g_{k,i} \sim \mathcal{N}(m_{t,i},\; v_{t,i}-m_{t,i}^2),
\end{equation}
where $m_{t,i}$ and $v_{t,i}$ denote the first and second moments of the gradient estimator.
We define the alignment probability as
\begin{equation}
\rho_i
=
\mathbb{P}\big[\text{sign}(-g_{k,i}) = \text{sign}((\Delta_k^r)_i)\big].
\end{equation}

If $(\Delta_k^r)_i>0$, alignment requires $-g_{k,i}>0$, i.e., $g_{k,i}<0$.
Thus,
\begin{equation}
\rho_i
=
\mathbb{P}[g_{k,i}<0]
=
\Phi\!\left(
-\frac{m_{t,i}}{\sqrt{v_{t,i}-m_{t,i}^2}}
\right),
\end{equation}
where $\Phi(\cdot)$ denotes the standard Gaussian CDF.
Similarly, if $(\Delta_k^r)_i<0$, alignment requires $g_{k,i}>0$, yielding
\begin{equation}
\rho_i
=
1-
\Phi\!\left(
-\frac{m_{t,i}}{\sqrt{v_{t,i}-m_{t,i}^2}}
\right).
\end{equation}

These cases can be unified as
\begin{equation}
\label{eq:rho-erf}
\rho_i
=
\frac{1}{2}
-
\frac{1}{2}
\,
\mathrm{erf}
\!\left(
\frac{m_{t,i}}{\sqrt{2(v_{t,i}-m_{t,i}^2)+\epsilon}}
\right)
\,
\text{sign}\big((\Delta_k^r)_i\big),
\end{equation}
where $\epsilon>0$ is a small constant for numerical stability. We define the correction scaling factor as
\begin{equation}
\gamma_{t,i}
=
1-\rho_i
=
\frac{1}{2}
+
\frac{1}{2}
\,
\mathrm{erf}
\!\left(
\frac{m_{t,i}}{\sqrt{2(v_{t,i}-m_{t,i}^2)+\epsilon}}
\right)
\,
\text{sign}\big((\Delta_k^r)_i\big),
\end{equation}
which corresponds to the implementation in Algorithm~\ref{alg:pflalign}:
\begin{equation}
\gamma_t
\leftarrow
0.5
-
0.5
\,
\mathrm{erf}
\!\left(
\frac{|m_t|}{\sqrt{2(v_t-m_t^2)+\epsilon}}
\right)
\text{sign}(-m_t \cdot \Delta_k^r).
\end{equation}

This formulation shows that $\gamma_{t,i}$ smoothly interpolates between full correction and minimal correction depending on the likelihood that the stochastic update direction disagrees with the accumulated personalized offset.
Importantly, the scaling relies only on local gradient moments and the personalized offset, avoiding explicit covariance estimation while retaining variance-aware control of aggregation-induced shrinkage.


\end{document}